\documentclass[letterpaper, 10 pt, conference]{ieeeconf}
\usepackage{cite}
\usepackage{amsmath,amssymb,amsfonts} 
\usepackage{graphicx}
\usepackage{textcomp} 
\usepackage{multirow}
\usepackage{gensymb}
\usepackage{xcolor} 
\usepackage[ruled,vlined]{algorithm2e}
\usepackage{gensymb}
\usepackage{textcomp} 
\usepackage{xcolor} 
\usepackage{color}
\usepackage{multirow}
\usepackage{graphicx}
\usepackage{subfigure}
\usepackage{caption}
\usepackage{booktabs} 
\usepackage[mathscr]{eucal}

\IEEEoverridecommandlockouts                              

\makeatletter

\makeatletter
\renewcommand{\maketag@@@}[1]{\hbox{\m@th\normalsize\normalfont#1}}%

\setbox0\hbox{$\xdef\scriptratio{\strip@pt\dimexpr
    \numexpr(\sf@size*65536)/\f@size sp}$}

\newcommand{\scriptveryshortarrow}[1][3pt]{{%
    \vcenter{\hbox{\rule[\scriptratio\dimexpr-.2pt\relax]
               {\scriptratio\dimexpr#1\relax}{\scriptratio\dimexpr.4pt\relax}}}%
   \mkern-4mu\hbox{\let\f@size\sf@size\usefont{U}{lasy}{m}{n}\symbol{41}}}}

\makeatother

\title{\LARGE \bf
Joint Camera Intrinsic and LiDAR-Camera Extrinsic Calibration 
}

\author{Guohang Yan, Feiyu He, Chunlei Shi, Pengjin Wei, Xinyu Cai and Yikang Li$^{\dagger}$ 
\thanks{$^{\dagger}$ Corresponding author.}
\thanks{Guohang Yan, Feiyu He, Chunlei Shi, Pengjin Wei, Xinyu Cai and Yikang Li are with Autonomous Driving Group, Shanghai AI Laboratory, China. {\tt\small \{yanguohang, hefeiyu, shichunlei, weipengjin, caixinyu, liyikang\}@pjlab.org.cn}}
}

\begin{document}
 
\maketitle

\begin{abstract}
Sensor-based environmental perception is a crucial step for autonomous driving systems, for which an accurate calibration between multiple sensors plays a critical role. For the calibration of LiDAR and camera, the existing method is generally to calibrate the intrinsic of the camera first and then calibrate the extrinsic of the LiDAR and camera. If the camera's intrinsic is not calibrated correctly in the first stage, it is not easy to calibrate the LiDAR-camera extrinsic accurately. Due to the complex internal structure of the camera and the lack of an effective quantitative evaluation method for the camera's intrinsic calibration, in the actual calibration, the accuracy of extrinsic parameter calibration is often reduced due to the tiny error of the camera's intrinsic parameters. To this end, we propose a novel target-based joint calibration method of the camera intrinsic and LiDAR-camera extrinsic parameters. Firstly, we design a novel calibration board pattern, adding four circular holes around the checkerboard for locating the LiDAR pose. Subsequently, a cost function defined under the reprojection constraints of the checkerboard and circular holes features is designed to solve the camera's intrinsic parameters, distortion factor, and LiDAR-camera extrinsic parameter. In the end, quantitative and qualitative experiments are conducted in actual and simulated environments, and the result shows the proposed method can achieve accuracy and robust performance. The open-source code is available at https://github.com/OpenCalib/JointCalib.

%
\end{abstract}

\section{INTRODUCTION}
Autonomous driving has attracted more and more attention in both industry and academic fields. 
To ensure that autonomous driving vehicles can operate appropriately under dynamic circumstances, multi-sensor cooperation technologies have appeared, which can provide adequate environmental information, thus contributing to a more reliable data fusion. Among the numerous types of sensor data fusion methods, the combination of LiDAR and camera is one of the most commonly used pairs of sensors for driving environment perception. LiDARs can provide 3D point cloud data, which include accurate depth and reflection intensity information, while cameras capture the rich semantic information of the scene. 
The combination of camera and LiDAR provides the feasibility to overcome the flaws of each sensor. The main challenge in fusing these two heterogeneous sensors is to find the precise camera's intrinsic parameters and the rigid body transformation between sensor coordinate systems by performing extrinsic calibration \cite{Guindel2017}. Researchers have made a lot of effort to improve the accuracy and efficiency of calibration results, such as specific targets like checkerboards \cite{cai2020,An2020,wang2017,liu2018,chen2021}, spherical target \cite{K2018}, gray code \cite{sels2019}, multi-plane stereo target \cite{Zhangjin2019} and semantic objects \cite{zhu2020}.
\begin{figure}[ht]
\centering
\includegraphics[scale=0.23]{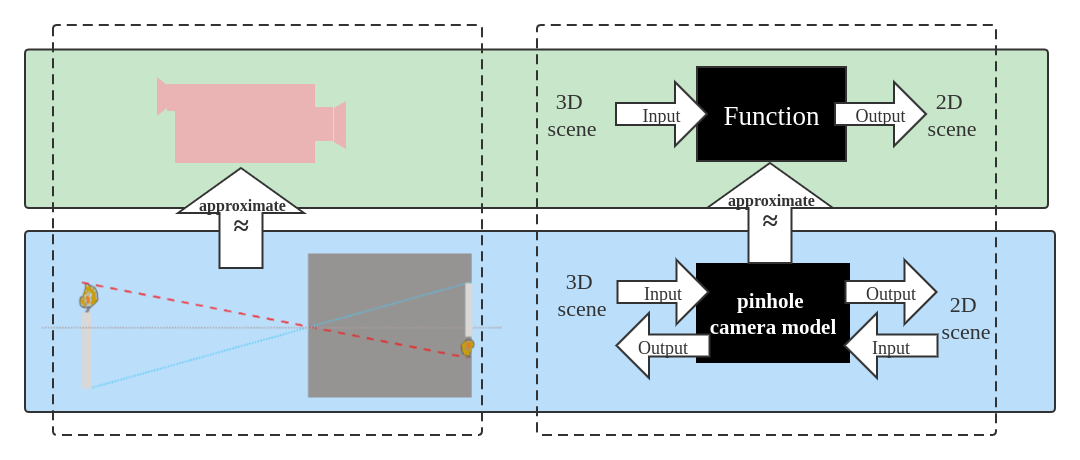}
\caption{The process of calibrating the camera's intrinsic parameters through a pinhole model.} \label{fig:pinhole}
\end{figure}

However, existing calibration methods suffer from various problems. The first issue is the reliability of the camera's intrinsic parameters because most methods assume the camera's intrinsic parameters are known or compute them through Zhang's process \cite{zhang2000}. The pinhole model is usually used when calibrating the camera's intrinsic, 
but the actual camera projection process and the pinhole model are not completely corresponding \cite{stu2006img}. The actual camera lens group is more complex and does not have an absolute optical center point \cite{juarez2020distorted}. As shown in Fig.~\ref{fig:pinhole}, the camera intrinsic calibration process is a pinhole model approximation measure. At the same time, due to the defects of the camera structure and the uncertainty in the optimization of nonlinear functions, the obtained solution is usually suboptimal. Consequently, the accuracy of extrinsic calibration will be affected. 
Our experiment of the camera's intrinsic calibration consistency also demonstrates the camera's intrinsic calibration volatility.

\begin{figure}[ht]
\centering
\includegraphics[scale=0.228]{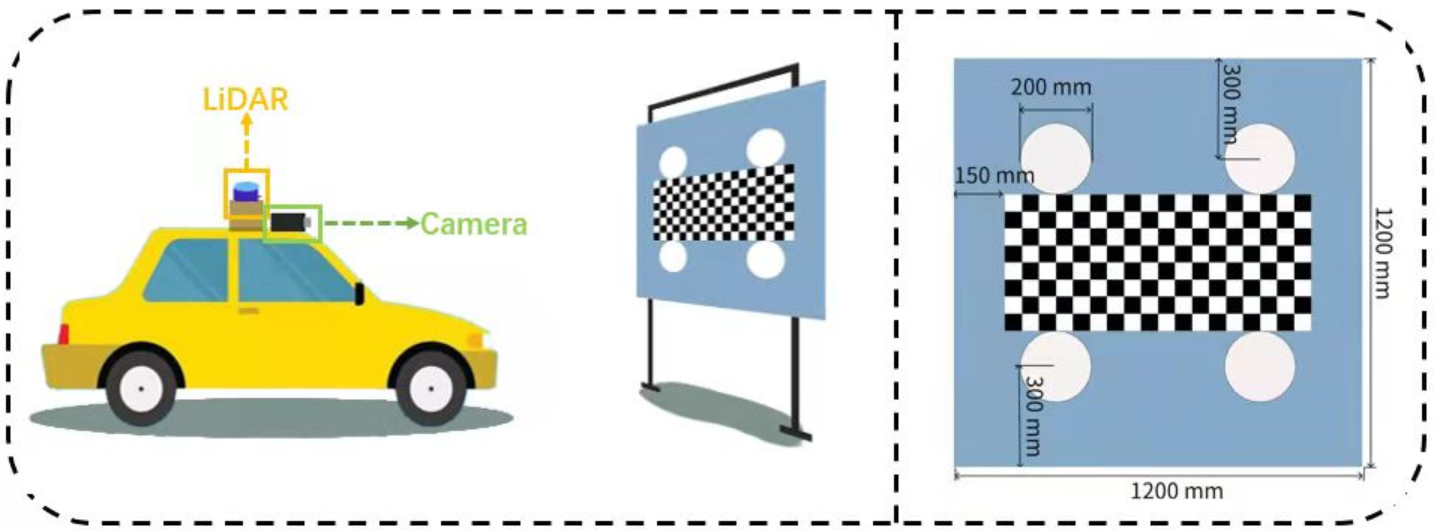}
\caption{A novel LiDAR-camera calibration board pattern.} \label{fig:board}
\end{figure}

In this paper, we present a novel joint calibration method that overcomes the inaccurate extrinsic calibration of LiDAR and camera caused by imperfect camera intrinsic parameters. Unlike existing calibration methods that only estimate the rotation and transformation between two sensor frames, the output of the proposed method contains the camera's intrinsic parameters, distortion factor, and LiDAR-camera extrinsic parameters. 

Firstly, as shown in Fig.~\ref{fig:board}, we design a novel calibration board pattern, which contains a checkerboard used for the calibration of the camera's intrinsic parameters and several circular holes for locating the LiDAR point cloud.
We first calibrate the camera initial intrinsic and board-camera initial extrinsic parameter by Zhang’s method \cite{zhang2000}. Then, 2D circles center points on the image are calculated from these parameters and the calibration board size. By extracting the position of the circles center in LiDAR, we can project the circles center 3D points to the image plane by the LiDAR-camera calibration parameters. The calculated 2D points and the projected 2D points form multiple 2D point pairs. We use the euclidean distance between these point pairs to refine the calibration parameters. At the same time, the constraints on reprojection of 3D-2D points of checkerboard corners are added to the optimization process.
%

The contributions of this work is listed as follows: 
\begin{enumerate}
\item We design a novel calibration board pattern that not only constrains the camera's intrinsic but can also be used to align LiDAR point clouds and camera images.

\item We propose a target-based joint calibration method based on our designed calibration board for the camera intrinsic and LiDAR-camera extrinsic parameters.
%

\item The joint calibration of the LiDAR-camera is formulated as a nonlinear optimization function by minimizing the reprojection error between the 3D-2D point pairs of circles center and checkerboard corners on the calibration board.

\item 
The proposed method shows promising performance on our simulated and real-world data sets; meanwhile, the related codes have been open-sourced to benefit the community.
\end{enumerate}

\begin{figure*}[ht]
\centering
\includegraphics[scale=0.18]{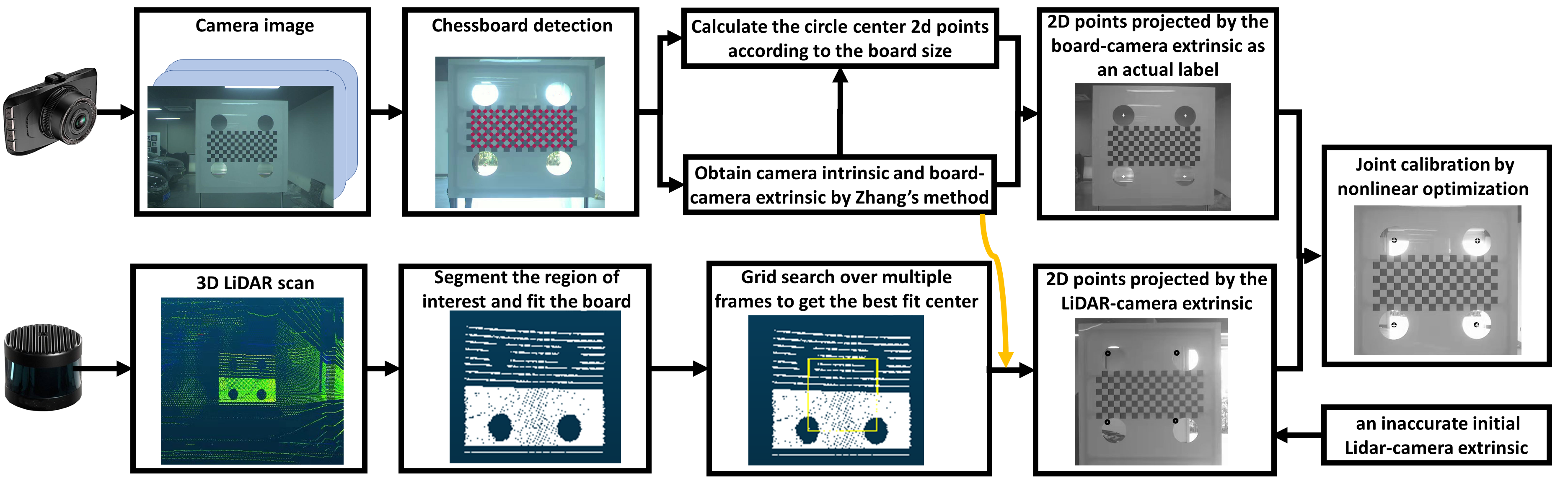}
\caption{Overview of the different stages of the presented method. First, target detection is performed, and then the camera intrinsic parameters and the board-camera extrinsic parameters are calculated by checkerboard to obtain the 3D-2D corresponding points of LiDAR and camera. Finally, a nonlinear optimization is performed to obtain the final calibration parameters. The inaccurate initial LiDAR-camera extrinsic parameters can be set by the sensor coordinate correspondence.} \label{fig:overview}
\end{figure*}

\section{RELATED WORK}
Researchers have proposed many approaches to address the intrinsic and extrinsic calibration of multi-model sensor calibration.
Intrinsic calibration estimates the operational parameters of sensors and is usually performed before performing the extrinsic calibration. Typically, camera intrinsic calibration focus on estimating focal length, distortion factor, and skewness. Escalera et al. \cite{Escalera2010} find both camera's intrinsic and extrinsic parameters by detecting corner points and extracting horizontal and vertical sets of lines from a checkerboard. Bogdan et al. \cite{bogdan2018deepcalib} estimate the focal length and distortion factor by a learning-based approach, and three different network architectures are proposed. Jin et al. \cite{jin2014} calibrate the intrinsic parameters of depth camera with cuboids, then optimize an objective function based on the distance error and angle error from reference cuboids.
An et al. \cite{an2018} apply Charuco board to overcome the deflects of checkerboard and ArUco board by building a Charuco board-based cube structure for feature points extraction, which can be used to estimate perspective projection matrix and solve the intrinsic parameters.
Lopez et al. \cite{Lopez_2019_CVPR} predict the extrinsic and intrinsic camera parameters through a single image by training a convolutional neural network. 
By contrast, extrinsic calibration estimates the rigid body transform between different sensor frames \cite{khaleghi2013}. According to the requirement of auxiliary equipment, extrinsic calibration can be divided into two categories: target-based and target-less procedures. 
\subsection{Target-based Method}
Target-based extrinsic calibration methods are widely used in the sensors calibration procedure. Researchers have designed various kinds of calibration targets to meet the characteristics of different sensors. 
Zhang et al. \cite{zhang2004} solve the extrinsic parameters based on a checkerboard and refine them by minimizing the re-projection error of laser points to the checkerboard plane. The method is easy to implement but cannot obtain the optimal solution directly.
Geiger et al. \cite{geiger2012automatic} propose an approach to calibrate LiDAR to the camera by extracting the corner points from both point cloud and image, then estimate the rotation and translation by maximizing the alignments of normal vectors and minimizing point-to-plane distances respectively. Finally, the fine registration is performed based on gradient descent. 
Huang et al. \cite{huang2020} perform extrinsic calibration by extracting vertices of the target from point cloud and image plane, and optimize the result by formulating a Perspective-n-Points problem which minimizes Euclidean distance of the corresponding corners. Intersection over Union (IoU) is also used for further refinement.
Zhou et al. \cite{zhou2018automatic} find the extrinsic parameters by computing the correspondences of line features in both LiDAR and camera frame, and refine the initial solution through a nonlinear optimization problem. 

\subsection{Target-less Method}
Target-less method usually takes advantage of natural environmental features (such as lines) from the scene, by solving the geometric constraints, the extrinsic parameters can be derived. 
Levinson et al. \cite{levinson2013} propose an online extrinsic calibration method by measuring the edge alignment. Edges in LiDAR points and image are extracted by depth discontinuity and Inverse Distance Transform (IDT) respectively, then minimize the re-projection error of 3D LiDAR edge points to the edge image to obtain the optimal solution.
Ma et al. \cite{ma2021crlf} calibrate the camera to LiDAR extrinsic by extracting line features in road scenes and registration via extracted road features.
Pandey et al. \cite{pandey2015automatic} use the reflection intensity measured by LIDAR and intensity values from camera and derive the optimal extrinsic parameters by maximizing a objective function based on mutual information.
Similarly, Taylor et al. \cite{Taylor2012AutomaticCO} perform LiDAR-camera calibration by normalizing the mutual information, and maximizing the gradient correlation of image and LiDAR points. In recent years, there are some works \cite{schneider2017regnet, shi2020calibrcnn, lv2021lccnet, zhao2021calibdnn} that apply deep learning to sensor calibration tasks, especially adapting to camera and LiDAR calibration problems. 

However, most methods assume that an accurate intrinsic parameter is given before performing the calibration procedure without paying attention to the problem that the inaccurate intrinsic will lead to inaccurate extrinsic calibration. 
In contrast, we provide a joint intrinsic and extrinsic calibration approach that outputs intrinsic and extrinsic parameters at the same time.

\section{METHODOLOGY}
This section introduces the details of our approach, including calibration target design, calibration target detection, calibration data collection, and calibration optimization process. Fig.~\ref{fig:overview} shows the overview of the presented method.

\subsection{Calibration target design}
According to the previous introduction, it is not easy to accurately calibrate the camera intrinsic parameters in practice. The calibration goal is to align the LiDAR point cloud with the camera image through intrinsic and extrinsic parameters. As long as the alignment accuracy of the point cloud and image is higher, the calibration's intrinsic and extrinsic parameters can be considered more accurate. 
So we designed a joint optimization algorithm, and the goal of joint optimization is to increase the alignment accuracy of the LiDAR point cloud and image. To this end, we need to design a novel calibration board that can achieve the joint optimization of intrinsic and extrinsic parameters.

A well-designed calibration target should satisfy the following properties (i) detectable in all relevant sensors and (ii) observable features extracted for localization. Based on the working principles of LiDAR and camera, depth discontinuous such as edges for point cloud and corners in the image is most widely used as calibration features since they can be detected accurately and robustly. 
As discussed in \cite{Velas2014CalibrationOR}, circular-shaped targets are more robust to be detected than rectangular shapes because they can interact with several LiDAR scans horizontally and vertically without missing edge information. As shown in Fig.~\ref{fig:board}, our design of calibration target combined both geometrical and visual characteristics, which is suitable for keypoints detection in LiDAR and camera modalities.
We put one black and white checkerboard in the center, surrounded by four circular holes.
On the one hand, the holes can use geometrical discontinuities in LiDAR point clouds. On the other hand, the checkerboard corners can provide spatial constraints. Fig.~\ref{fig:board} also shows the details of our calibration target with a specific size. It is worth mentioning that the calibration board pattern can also be further modified based on this design to make data collection more convenient.

\subsection{Calibration target detection}
The first step of target-based calibration is to locate the spatial position of the calibration in each sensor frame. LiDAR sensor returns the 3D position while camera returns the colored 2D image of the target. Here we conduct the extraction procedures of checkerboard and circular holes separately.
There are many existing techniques for detecting the checkerboard from the image, for example, using popular computer vision libraries. In this work, we choose OpenCV \cite{opencv_library} library for checkerboard detection. 
As for LiDAR data, different from the method mentioned in \cite{beltran2021automatic}, we locate each center of holes by generating a point cloud mask $\mathcal{P}_{mask}^L$, which has the same geometric structure as the calibration target.
\begin{figure}[ht]
\centering
\includegraphics[scale=0.4]{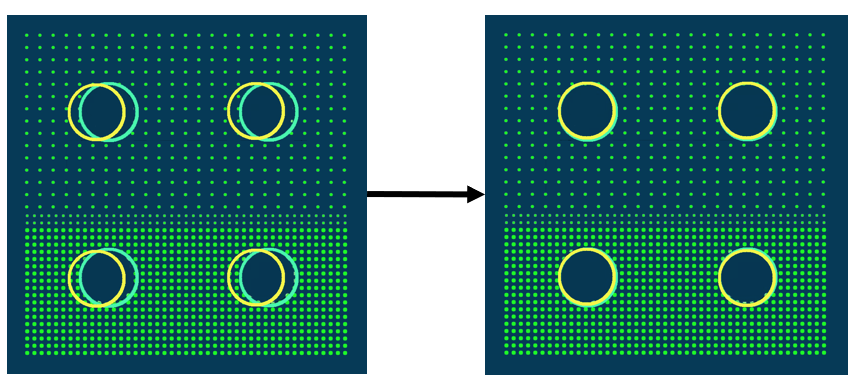}
\caption{The process of accurately extracting the center of the circle on the calibration board.} \label{Fig:circle}
\end{figure}
We assume the original point cloud data scanned by a LiDAR sensor is represented as $\mathcal{P}_0^L$. Firstly, in order to filter outliers and remove the noise of the surrounding environment, we segment the regions of interest by presetting detection range.
Then the calibration target plane is segmented from $\mathcal{P}_0^L$ through RANSAC plane fitting with orientation constraints, which is denoted by $\mathcal{P}_{target}^L$.
Afterwards, we follow the grid search method described in \cite{levinson2013} to find the best match of $\mathcal{P}_{mask}^L$ and $\mathcal{P}_{target}^L$ in LiDAR coordinate system \{L\}. Rather than searching on 6-DoF, here we focus on the yaw angle for rotation and $x,y$ for translation relative to the calibration board plane.
\begin{equation}\label{simplified cost function}
\begin{aligned}
R_{yaw}^*,t_{x, y}^* = \arg\underset{yaw,x,y}\min \sum(R_{yaw}(\mathcal{P}_{mask}^L + t_{x, y}) \cap \mathcal{P}_{target}^L) 
\end{aligned} 
\end{equation}
Fig.~\ref{Fig:circle} illustrates the principle of the matching procedure, when perfect alignment achieved, least number of points will fall into the circular holes, and thus the 3D position of each hole center can be located in \{L\}.
Compared with \cite{beltran2021automatic}, our method only requires the coordinate of the 3D point cloud without using additional intensity and ring id information, and the generated mask fixes the relative position between holes, which greatly improves the robustness and accuracy of center detection. Finally, each calibration board can obtain the center points of the four circular holes in the LiDAR coordinate system.

The calibration data collection process is similar to the camera’s intrinsic calibration data collection. As shown in Fig.~\ref{Fig:collection}, in order to ensure the calibration accuracy, checkerboard corner features and circular hole features need to be collected in each area of the image. In practical applications, multiple calibration boards can be placed in different pose positions simultaneously so that we can perform calibration by collecting only a frame of data. Compared to the targetless calibration method, this method guarantees the alignment of point clouds and pixels in all image regions, ensuring that the near feature alignment is accurate enough to ensure far alignment accuracy. This solution is mainly used in the calibration room or production line calibration. After the calibration board environment is set and fixed, it can be used for LiDAR-to-camera intrinsic and extrinsic parameter calibration of a large number of autonomous vehicles.

\begin{figure}[ht]
\centering
\includegraphics[scale=0.28]{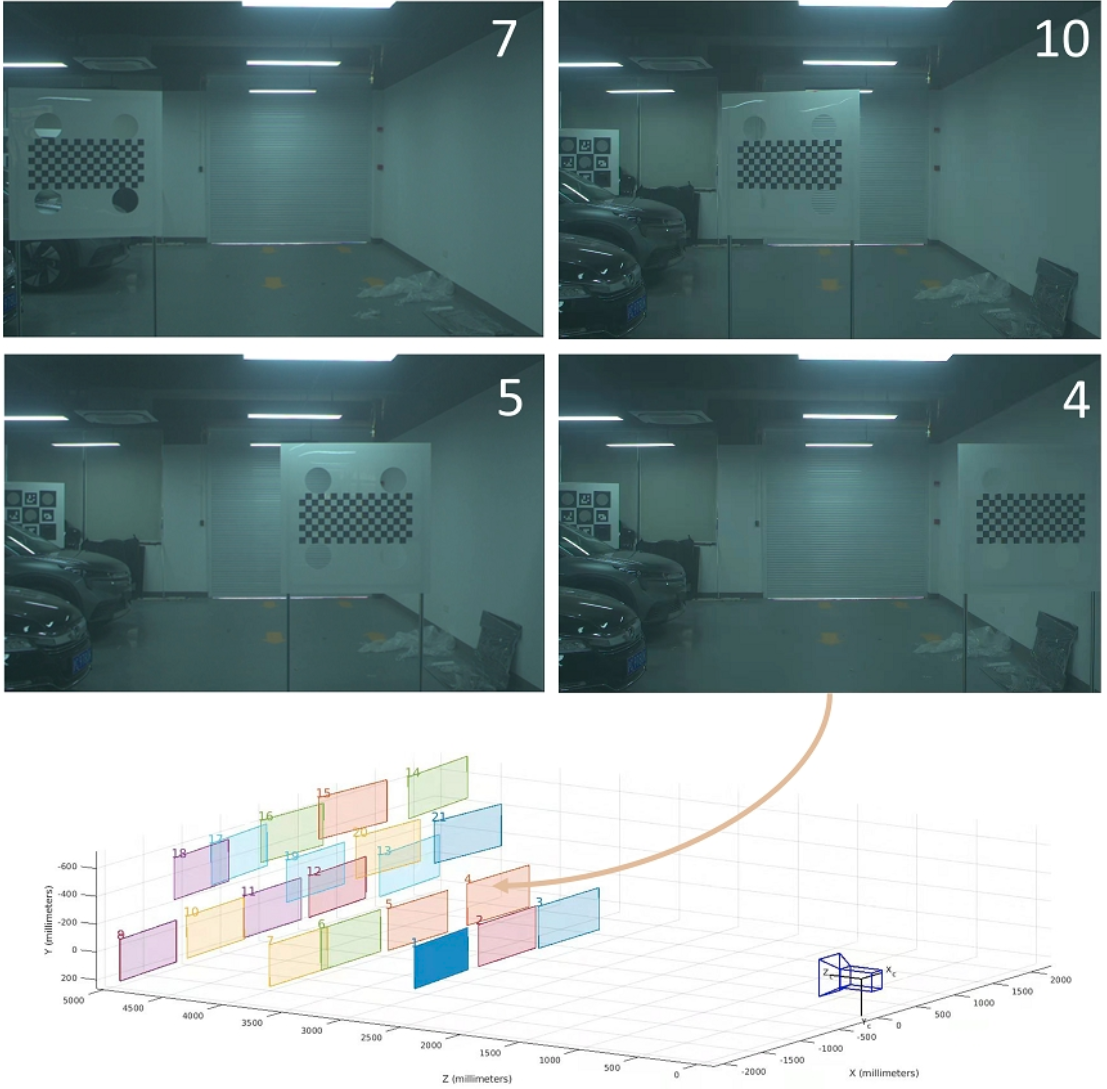}
\caption{Collect the point cloud and checkerboard data of the main area in the image.} \label{Fig:collection}
\end{figure}

\subsection{Joint optimization process}
\subsubsection{Circle Center 2D Points Calculation}


For the extracted LiDAR 3D circles center points, we need to get the corresponding 2D points on the image.
We first calibrate the camera intrinsic $\mathcal{K}$, distortion factor and board-camera extrinsic $\textbf{T}(t_x,t_y,t_z,r_x,r_y,r_z)$ by checkerboard corners. According to the size of the calibration board, we get the four circles center points  $P_B$$=$$\{p_b^1, p_b^2, \cdots , p_b^4\}$$\in$$\mathbb{R}^3$ of each calibration board, where $B$ represents the coordinate system of the calibration board. The transformed point $\mathbf{p}_c^\mathrm{i}$ is then projected onto the camera image plane by $\mathcal{K}$ and $\textbf{T}$.
\begin{equation}
\label{equ:extrinsic}
    \mathbf{p}_c^\mathrm{i} = \mathcal{K}(\boldsymbol{\mathrm{R}}(\mathbf{r}) \cdot \mathbf{p}_b^\mathrm{i} + \boldsymbol{\mathrm{t}}), \quad k=1,2,\cdots,4.
\vspace{-2mm}
\end{equation}
where $\boldsymbol{\mathrm{R}}(\textbf{r})$ represents the rotation and can be parameterized by angle-axis representation $\mathbf{r}$. $\boldsymbol{\mathrm{t}}$ represents the translation with $\boldsymbol{\mathrm{t}}$$=$$(t_x,t_y,t_z)^T$. After adding distortion factor, the actual location of the projected point $p_i$=$(u_i,v_i)$ is
\vspace{-2mm}
\begin{equation}
\label{equ:extrinsic}
    \mathbf{p_i} =\boldsymbol{\mathrm{D}}(\mathbf{p}_c^\mathrm{i}),
\vspace{-2mm}
\end{equation}
$ \boldsymbol{\mathrm{D}}(p)$ is the camera distortion model. The extracted LiDAR 3D circles center points and the calculated 2d circles center points form multiple 3D-2D point pairs.

\subsubsection{Objective}
Through the previous process, multiple sets of 3D-2D point pairs $\boldsymbol{S}$ of the circles center 
on the calibration board are obtained. The calibration goal is to align the LiDAR point cloud with the camera image through intrinsic and extrinsic parameters. Therefore, to align the point cloud with the image, we minimize the following objective function:
\begin{equation}
    \boldsymbol{E} = \underset{i\in \boldsymbol{S}}\sum || \boldsymbol{\mathrm{D}}(\mathcal{K}(\boldsymbol{\mathrm{R}_{LC}} \cdot \mathbf{p}_{3D}^\mathrm{i} + \boldsymbol{\mathrm{t}_{LC}})) -\mathbf{p}_{2D}^\mathrm{i} ||^2  \ 
\end{equation}
where $\boldsymbol{\mathrm{R}_{LC}}$, $\boldsymbol{\mathrm{t}_{LC}}$ represents the extrinsic parameter from LiDAR to the camera. $\mathbf{p}_{2D}$ are the pixel coordinates of the center of the circles calculated by the checkerboard before joint optimization. $\mathbf{p}_{3D}$ are the 3D points of the circles in the LiDAR coordinate system.

\subsubsection{Constraints}
The above minimization is subject to a set of constraints. While ensuring the alignment of the LiDAR and the image, the constraints of the checkerboard on the camera's intrinsic parameters are also required. the 3D points $P_{corners}$$\in$$\mathbb{R}^3$ of the chessboard corners are in the coordinate system \{B\} of the calibration board.
For the camera's intrinsic parameters, the constraints are listed below:
\begin{equation}
\begin{aligned}
     \sum\limits_{(u,v) \propto P_{corners}}(||u - u_{det}||^2 + ||v - v_{det}||^2) &= 0 
\end{aligned}
\label{}
\end{equation}
where $(u,v)$ represents the pixel coordinate point of the $P_{corners}$ projection, $(u_{det}, v_{det})$ is the actual detected chessboard corner pixel point. The rest of the constraints are that the 2D position of the circles center calculated by the calibration board size needs to remain invariant during optimization. The points set $\mathbf{b}^\mathrm{i} = [X_i, Y_i, 0]$ are the circles center in the calibration board coordinate system.
\begin{equation}
    \underset{i\in \boldsymbol{S}}\sum || \boldsymbol{\mathrm{D}}(\mathcal{K}(\boldsymbol{\mathrm{R}_{BC}} \cdot \mathbf{b}^\mathrm{i} + \boldsymbol{\mathrm{t}_{BC}})) -\mathbf{p}_{2D}^\mathrm{i} ||^2 = 0  \ 
\end{equation}
where $\boldsymbol{\mathrm{R}_{BC}}$, $\boldsymbol{\mathrm{t}_{BC}}$ represents the extrinsic parameter from the calibration board to the camera. 

\subsubsection{Nonlinear Solution}
The rotation matrix uses nine vectors to describe the rotation of 3 degrees of freedom, which is redundancy. Furthermore, the rotation matrix has to be an orthogonal matrix with determinant 1. These constraints increase the difficulty of the solution when estimating or optimizing rotation matrices. A better way to represent the rotation matrix is using the angle-axis rotation vector. In our implementation, an inaccurate initial Lidar-camera extrinsic are provided as initialization, and the calibration optimization equation is solved by the Ceres solver\cite{ceres-solver}.
\section{EXPERIMENTS}
\vspace{-1mm}
The experiment in this paper consists of two parts: a realistic experiment on our driverless vehicle test platform and a simulated experiment based on the Carla engine \cite{Dosovitskiy17}. The results show that the proposed method is superior to the state-of-the-art in terms of accuracy and robustness.
\subsection{Realistic Experiment}
\begin{figure}[h]
\centering
\includegraphics[scale=0.28]{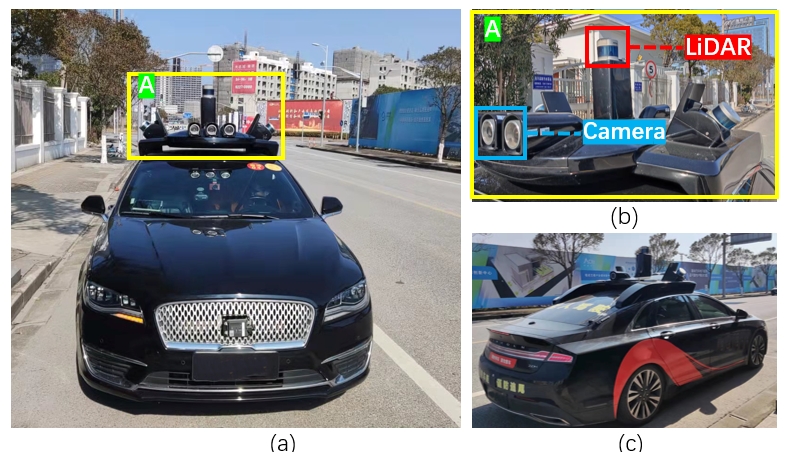}
\caption{Our sensor suit. Top is Hesai Pandar64 LiDAR. Front is the camera Balser acA1920-40gc with  different FOVs ($FOV=30^{\circ}$, $FOV=60^{\circ}$, $FOV=120^{\circ}$).} \label{Fig:realistic_setup}
\end{figure}

We conducted experiments on real driverless platforms, Fig.~\ref{Fig:realistic_setup} shows our realistic experiment setup.

\subsubsection{Camera Intrinsic Calibration Consistency Evaluation}
\begin{figure}[h]
\centering
\includegraphics[scale=0.23]{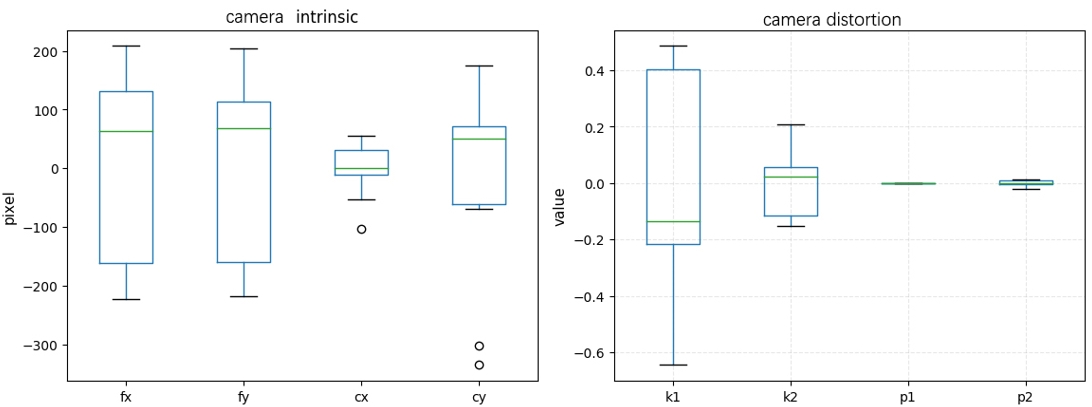}
\caption{Camera's intrinsic and distortion parameter calibration consistency evaluation.} \label{Fig:fov60_evaluation}
\end{figure}

Due to the complex internal structure of the camera and the way of data acquisition, the camera's intrinsic parameters are usually unstable in calibration \cite{stu2006img}. On the other hand, the inaccuracy of the camera's intrinsic parameters is because the actual camera projection process and the pinhole model are not completely corresponding, and the equivalent camera optical center points at different distances are different \cite{juarez2020distorted}. Barrel distortion usually occurs at short focal lengths, and pincushion distortion usually occurs at long focal lengths\cite{bukhari2013automatic}. To evaluate the camera's intrinsic calibration instability, We designed the camera intrinsic calibration consistency experiment.
%
We used the camera to collect 6 uniformly distributed checkerboard image groups, and each group contained 100 frames. Then, we randomly selected 25 images from each group for the camera's intrinsic calibration, performed 100 times per group. We obtained the residual vector whose statistical information (e.g. mean, variance) reveals the camera's intrinsic calibration volatility. The results are shown in Fig.~\ref{Fig:fov60_evaluation}. 

%

%


%
%

\subsubsection{Ablation experiment}
We designed an ablation experiment to evaluate our method better to split our one-stage approach into a two-stage process. Then, we respectively compared the 3D-2D reprojection error of the circles center and checkerboard corners on the calibration board. We conducted six sets of experiments, each was calibrated with one-stage and two-stage respectively and compared the reprojection error of their checkerboard and the circles center. As shown in Table \ref{table:reprojection}, our method has a smaller reprojection error of the circles center than the two-stage calibration.
\begin{table}[htbp]
\caption{Average Reprojection Error for One-stage and Two-stage Calibration}
\centering
\renewcommand{\arraystretch}{1.3}
\begin{tabular}{|c|c|c|c|c}
\hline
 & Checkerboard Corners & Circles Center \\
\hline
One-stage & $0.757$ pixel & $1.104$ pixel \\
\hline
Two-stage & $0.546$ pixel & $5.428$ pixel \\
\hline
\end{tabular}  
\label{table:reprojection} 
\end{table}

\subsubsection{Qualitative Results}
\begin{figure}[ht]
\centering
\includegraphics[scale=0.35]{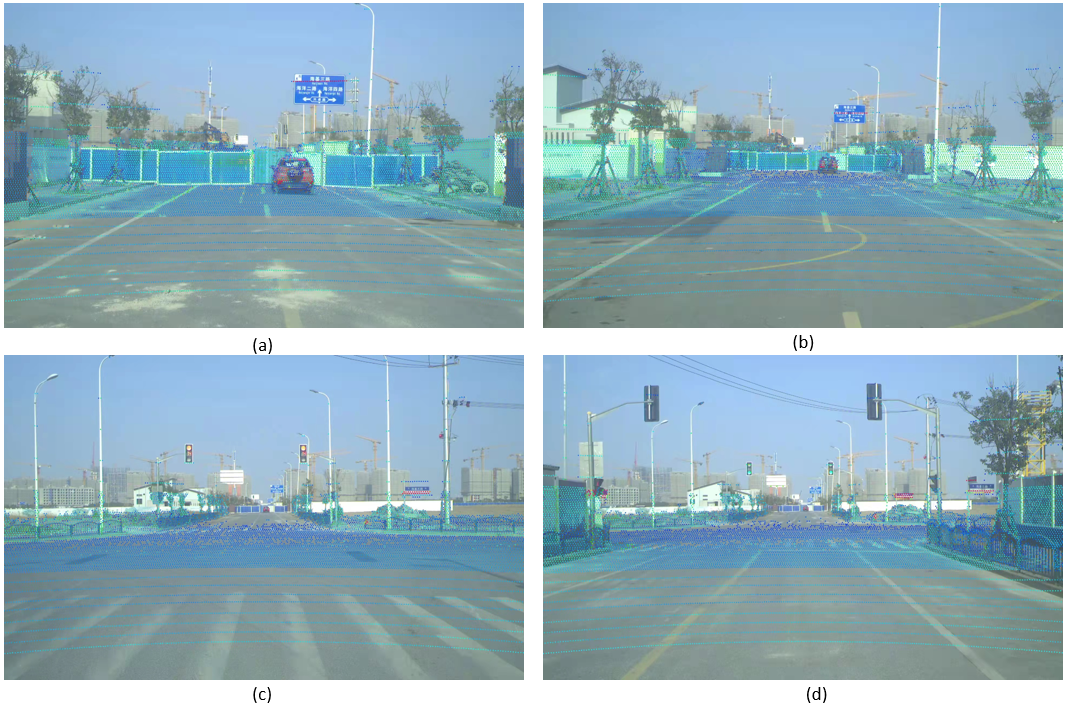}
\caption{Point cloud projections in four different scenarios, projection points color is represented by the LiDAR intensity.} \label{Fig:lidar-camera_projection}
\end{figure}

\begin{figure}[ht]
\centering
\includegraphics[scale=0.365]{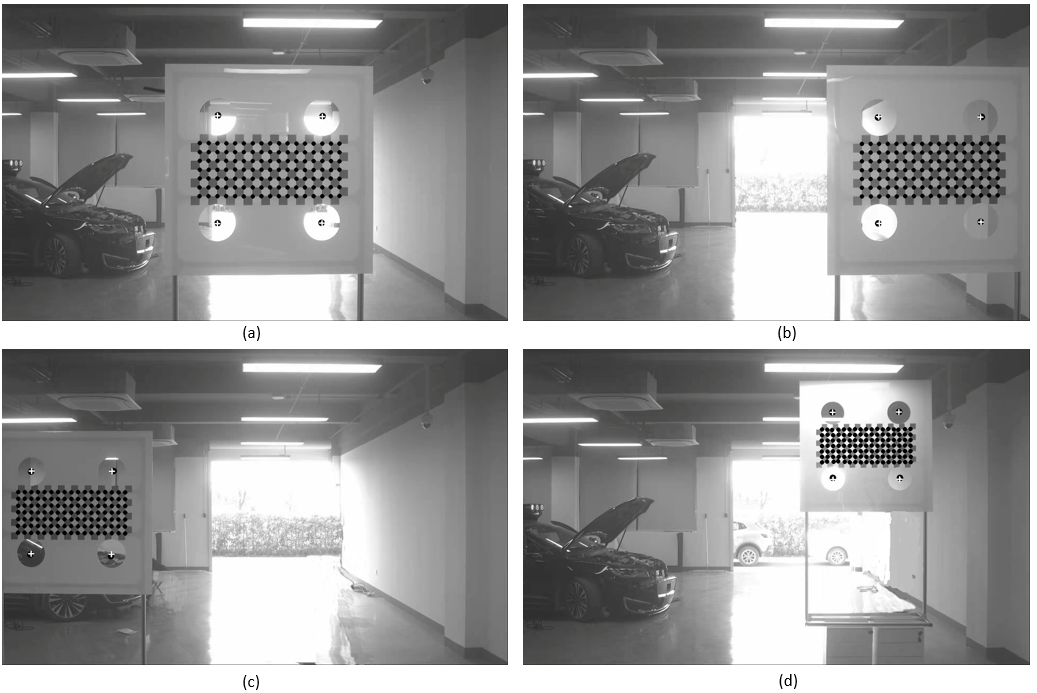}
\caption{Realistic Experiment: Reprojection of the circles center on the calibration board. The black circle is the LiDAR projection point, and the white cross is the calculated image point.} \label{Fig:circle_reprojection}
\end{figure}

To better visualize the performance of our method, the point cloud is projected to the image plane using the intrinsic and extrinsic parameters calibrated by our method. Results are shown in Fig.~\ref{Fig:lidar-camera_projection}. As shown, using the calibration parameters extracted by the proposed approach enables a perfect alignment between both data modalities. Fig.~\ref{Fig:circle_reprojection} shows the effect of circles center reprojection on the calibration board.

\begin{figure}[h]
\centering
\includegraphics[scale=0.35]{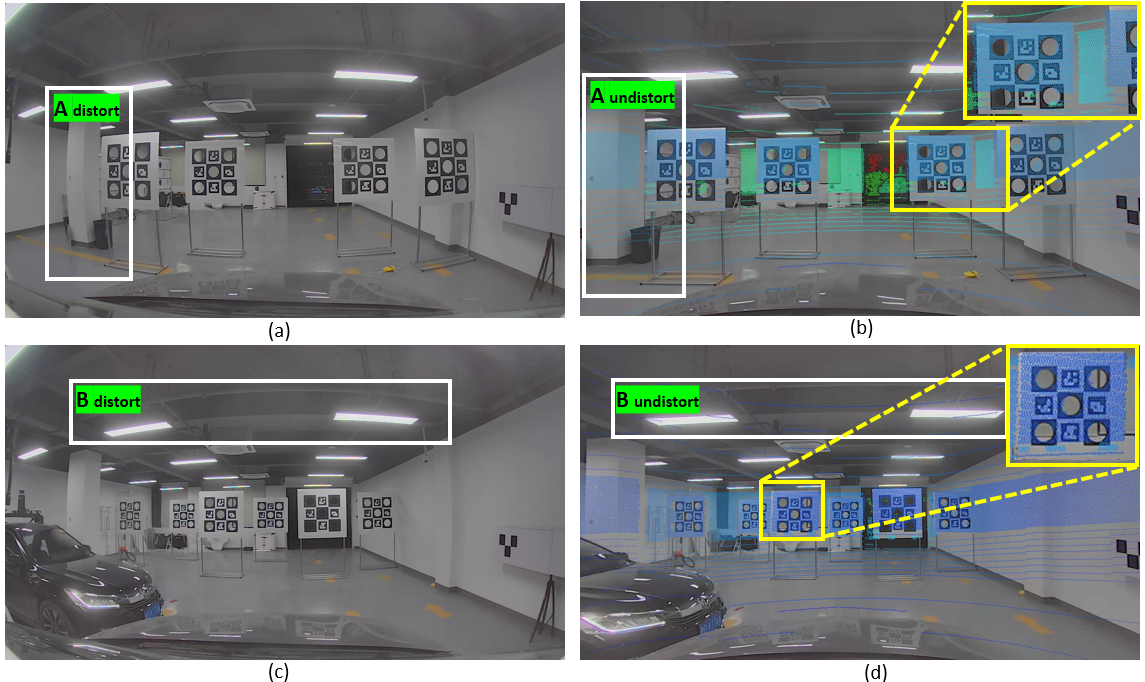}
\caption{LiDAR-camera extrinsic calibration by the method \cite{beltran2021automatic}. The small intrinsic error leads to a decrease in the extrinsic calibration accuracy.} \label{Fig:comparison}
\end{figure}

\subsubsection{Comparison Experiments}
We compared our methods with \cite{zhou2018automatic} and \cite{beltran2021automatic}, which both use checkerboard as a calibration target. The first one method \cite{zhou2018automatic} estimates the extrinsic by minimizing the distance from LiDAR points to the checkerboard plane estimated from the image. The second method \cite{beltran2021automatic} estimates the extrinsic by ArUco markers and circles center registration on the calibration board. We first spent a lot of effort calibrating the camera's intrinsic parameter. As shown in Fig.~\ref{Fig:comparison}, the A and B boxed in the image show that the distortion effect is good. However, due to the small error of the camera focal length and optical center, the parameters calibrated by the method \cite{zhou2018automatic} and \cite{beltran2021automatic} still cannot align the point cloud and the image in some places. For this case, our method can adjust the intrinsic parameters so that the point cloud and the image are perfectly aligned. 

\subsection{Simulated Experiment}
We also conducted experiments with our method in the simulation environment. We can get the ground truth of the sensor's calibration in the simulation environment. We generated three calibration data groups in the Carla engine \cite{Dosovitskiy17}, 
We quantitatively compared the mean of LiDAR-camera extrinsic calibration with the ground truth in Table \ref{table:result}. Fig.~\ref{Fig:comparison} also shows the effect of circles center reprojection on the calibration board. The result shows the proposed method can achieve accuracy and robustness performance.

\begin{figure}[h]
\centering
\includegraphics[scale=0.202]{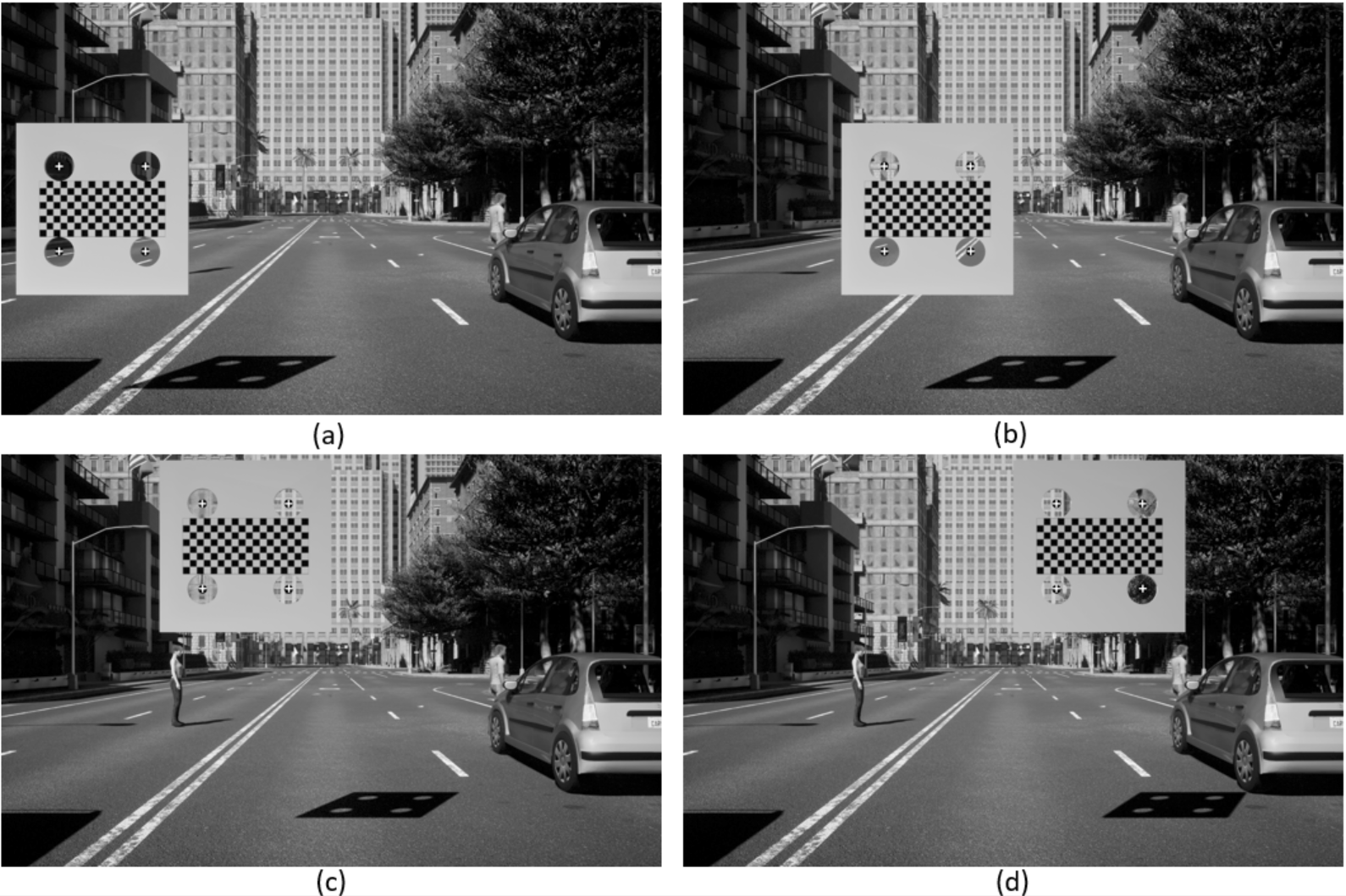}
\caption{Simulated Experiment: Reprojection of the circles center on the calibration board. The black circle is the LiDAR projection point, and the white cross is the calculated image point.} \label{Fig:collection}
\end{figure}
\begin{table}[tbp]
\centering
\caption{Translation and Rotation Quantitative Evaluation.} 
\renewcommand{\arraystretch}{1.3}
\begin{tabular}{|c|c|c|c|c|c|c|c|}
\hline
   & $ t_x$(m) & $ t_y$(m) & $ t_z$(m) & $$\textit{roll}(\degree) &  $$\textit{pitch}(\degree) &  $$\textit{yaw}(\degree) \\ \hline
GT & $0$ & $0.595$ & $2.5$ & $-90$ & $0$ & $90$    \\ \hline
Our & $-0.001$ & $0.5912$ & $2.5079$ & $-90.002$ & $0.011$ & $90.004$    \\ \hline
\end{tabular}  

\label{table:result}
\end{table}

\section{CONCLUSIONS}
This paper proposes a novel target-based calibration method, joint camera intrinsic and LiDAR-camera extrinsic calibration. The method can reduce the inaccurate LiDAR-camera extrinsic calibration caused by incorrect intrinsic parameters and can be applied in calibration rooms or factory calibration. Qualitative and quantitative results demonstrate the performance and effectiveness of our method. The related codes and data have been open-sourced to benefit the community. In the real environment, due to the sparsity of the point cloud, the extraction accuracy of the circle center may decrease. In the future, we look forward to using the multi-frame data of vehicle motion to increase the density of the point cloud, thereby improving the extraction accuracy of the circle center to improve the calibration performance further. Additionally, we will explore the detection and calibration accuracy of different types of LiDARs, such as repetitive scanning, non-repetitive scanning, sparse, dense, etc.

\bibliographystyle{IEEEtran}
\bibliography{egbib}

\end{document}